\documentclass[letterpaper, 10 pt, journal, twoside]{IEEEtran}

\usepackage{dsfont}
\usepackage{cite}
\usepackage{url}
\usepackage{comment}
\usepackage{algorithmic}
\usepackage{graphicx}
\usepackage{amsmath,amssymb,amsfonts}

\usepackage{multirow}

\newcommand{\ie}{{\em i.e.}}
\newcommand{\eg}{{\em e.g.}}

\usepackage{color}
\definecolor{red}{rgb}{1.00,0.20,0.20}
\definecolor{blue}{rgb}{0.20,0.20,1.00}
\definecolor{green}{rgb}{0.00,1.00,0.00}

\usepackage{array}
\newcolumntype{L}[1]{>{\raggedright\let\newline\\\arraybackslash\hspace{0pt}}m{#1}}
\newcolumntype{C}[1]{>{\centering\let\newline\\\arraybackslash\hspace{0pt}}m{#1}}
\newcolumntype{R}[1]{>{\raggedleft\let\newline\\\arraybackslash\hspace{0pt}}m{#1}}

\ifdefined\UseShortAcronyms

\else

\fi

\IEEEoverridecommandlockouts                              
\title{Attentional Graph Neural Network for Parking-slot Detection}

\author{Chen Min and Jiaolong Xu$^{*}$ and Liang Xiao and Dawei Zhao and Yiming Nie and Bin Dai
\thanks{$^{*}$ Corresponding author. Email: {\tt\small jiaolong\_xu@126.com}}%
}

\begin{document}

\maketitle

\begin{abstract}
Deep learning has recently demonstrated its promising performance for
vision-based parking-slot detection. However, very few existing methods explicitly take into account learning the link information of the marking-points, resulting in complex post-processing and erroneous detection. In this paper, we propose an attentional graph neural network based parking-slot detection method, which refers the marking-points in an around-view image as graph-structured data and utilize graph neural network to aggregate the neighboring information between marking-points. Without any manually designed post-processing, the proposed method is end-to-end trainable. Extensive experiments have been conducted on public benchmark dataset, where the proposed method achieves state-of-the-art accuracy. Code is publicly available at \url{https://github.com/Jiaolong/gcn-parking-slot}.
\end{abstract}

\section{Introduction}
\IEEEPARstart{A}{utonomous} valet parking (AVP) is an important application for autonomous vehicles, where parking-slot detection plays a critical role \cite{DeepPS}. Compared with manual parking, AVP can provide more accurate parking path and safer control, which can reduce the scratch and collision caused by manual operation errors. 

Recently, an increasing number of vehicles are equipped with around-view monitor (AVM) systems for better observation of the surrounding road conditions\cite{PS-DCNN}. Parking-slot detection with around-view image, namely vision-based parking-slot detection, has shown great potential for AVP system, as it can accurately identify the junctions and parking lines of the parking-slot \cite{DeepPS}. 

Vision-based parking-slot detection methods with hand-crafted features are firstly proposed\cite{2000-first, 2006-second, 2010-third}, but they are not robust under complex environmental conditions. In recent years, with the rise of deep learning, deep convolutional neural network (CNN) based parking-slot detection approaches have achieved good accuracies \cite{DeepPS, DMPR-PS}. Most of the deep learning based methods adopt multi-stage processing, {\eg}, the marking-points (the vertexes of the parking-slot) are firstly detected by convolution neural network, and parking-slots are predicted by post-processing using manually designed geometric rules \cite{DMPR-PS}. Although CNN-based parking-slot detection approaches provide promising results, they have two main drawbacks. Firstly, these methods detect the marking-points of parking-slot separately without considering the link information between them. Secondly, they need post-processing to obtain the parking-slot, which is time consuming and inaccurate.

The existing object detection methods usually assume the objects in an image are independent to each other. However, this is not suitable when the objects (nodes) in one image have relation with the others (neighbors), {\ie}, the graph-structured data. In an around-view image, one marking-point is naturally related to others, {\ie} forming a parking-slot or not, thus these marking-points are natural graph-structured data. Leveraging such link information between marking-points (nodes) can be beneficial to accurate parking-slot detection. The graph-structure property of marking points motivates us to use Graph Neural Networks (GNN) \cite{GCN} to model the relationships between marking-points. Inspired by the success of the Transformer \cite{Transformer}, we propose to incorporate attention mechanism in GNN for parking-slot detection. By utilizing attention mechanism, nodes on the graph are able to specify different weights to different nodes in a neighborhood, enabling implicitly learning link information between marking-points. Unlike previous CNN based methods \cite{DeepPS, DMPR-PS}, the proposed attentional GNN based parking-slot detection model integrates marking-points detection and parking-slot inference in a single stage model, {\ie}, end-to-end trainable.

\begin{figure*}[ht]
	\centering
	\centerline{\includegraphics[width=4in]{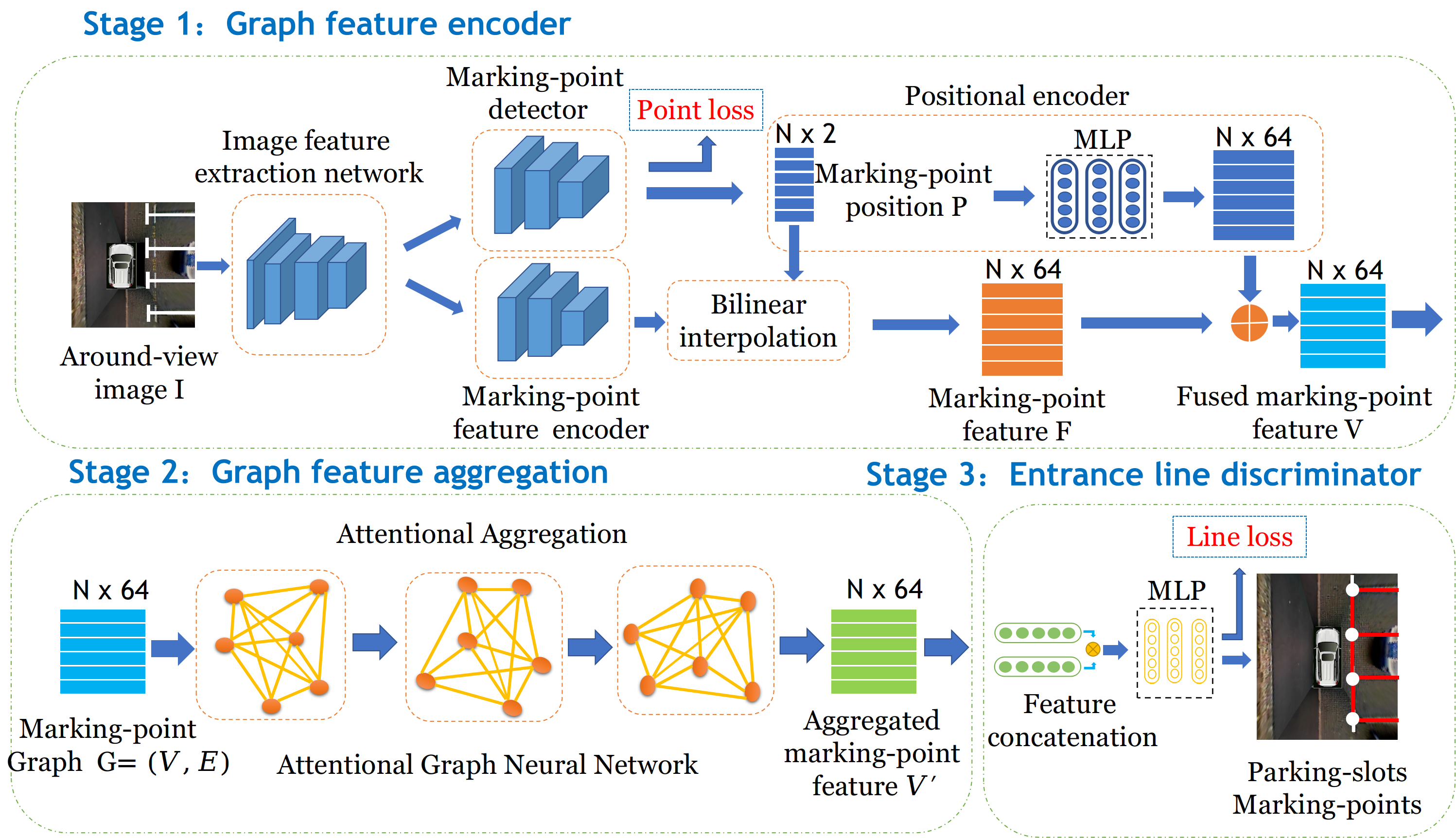}}
	\caption{The overall architecture of the proposed method. The around-view images are first fed into the image feature extraction network to extract the deep image features. The image features are then sent into the marking-point detector to detect marking-points and the marking-point encoder to extract marking-point features. The marking-point features and positions are then sent to the attentional GNN to learn neighboring information. Finally we apply an entrance line discriminator to decide whether the two marking-points form an entrance line.
	}
	\label{network_overall}
\end{figure*}

The overall architecture of proposed attentional GNN based parking-slot detection model is depicted on Fig.~\ref{network_overall}. The proposed method consists of three main components: graph feature encoder, graph feature aggregation and entrance line discriminator. Specifically, the marking-point detector detects marking-point first and then extract the deep features for each marking-point. The feature aggregation network are then used to fuse the position information and deep features and aggregate the neighboring information with attentional graph neural network. The encoded discriminative features of the marking-point pairs are processed by the entrance line discriminator network to decide whether or not they can form an entrance line. The highlights of our work are as follows:
\begin{itemize}
	\item We model the marking-points in the around-view image as graph-structured data, and design an attentional graph neural network to aggregate the neighboring information between marking-points to boost parking-slot detection performance.
	\item We propose an end-to-end trainable parking-slot detection method without any manually designed post-processing.	
	\item The proposed method does not require fine-grained annotations, such as direction and shape of marking-points, thus reduces the training expenses.
\end{itemize}

\section{Related Work}
\subsection{Traditional parking-slot detection}
Vision-based parking-slot detection has been extensively studied for decades \cite{DeepPS, DMPR-PS}. 
Traditional parking-slot detection methods can be categorized into line-based ones and marking-point-based ones. The line-based methods first detect lines in the around-view image via various techniques such as Canny edge detector \cite{randn-1} and Sobel filter \cite{sobal-1}, and then they predict the parameters of the detected lines via line fitting algorithms.
The manual designed geometric constraints are then applied to filter and locate the parking-slots. 
Similar to line-based methods, marking-point-based methods first find the marking-points in the around-view image via harris corner detector \cite{harris-1} or boosting decision tree \cite{2000-first}, and then use template matching technique \cite{harris-2} or combine line detection to locate the parking-slots \cite{line}. Although these traditional parking-slot detection methods generally provide good results, they are sensitive to environmental changes and not applicable to complex real-world environment.

\subsection{Deep learning-based parking-slot detection}

Recently, deep learning-based approaches have been proposed to directly solve the parking-slot detection problem. DeepPS \cite{DeepPS} introduces the first learning-based pipeline for parking-slot detection. It first utilizes a CNN network to detect the marking-points in the around-view image, and then uses another CNN network to match paired marking-points. DeepPS has achieved promising results under different environmental conditions. DMPR-PS \cite{DMPR-PS} is another two-stage method. In the first stage,  a novel CNN-based model is used to regress the orientation, coordinate, and shape of marking-points, and in the second stage, the manually designed geometric rules are applied to filter and match paired marking-points. Our method also uses CNN to regress the coordinates of marking-points for parking-slot detection. However, instead of using complex and inconvenient post-processing in DMPR-PS \cite{DMPR-PS}, we build a fully connected graph for the marking-points to explore the neighboring information and directly predict the parking-slots in an end-to-end way.

\subsection{Graph convolutional networks}
Graph Neural Networks (GNN) is initially introduced in \cite{gcn:2009} and has become popular for modeling global relations. GNN models can be categorized into spectral approaches \cite{bruna2013spectral, defferrard2016convolutional} and non-spectral approaches \cite{GCN, GAT}. Our work is related to \cite{GCN, GAT} which conduct spatial convolution by aggregating node features in local neighborhoods on graph. GNN models have been applied to many computer vision tasks such as human pose estimation \cite{gcn_pose_estimation}, image matching \cite{Superglue} and point cloud segmentation \cite{poitcloud-grpah}. In this work, we design attentional graph neural networks for parking-slot detection. As far as we know, this is the first work to apply GNN for parking-slot detection.

\section{Method}
\subsubsection{Overview of the proposed method}

Assume a parking-slot consists of four marking-points $(P_1, P_2, P_3, P_4)$, the parking-slot detection problem in this work is formulated as the problem of detecting an ordered marking-point pair $(P_1,P_2)$ on the entrance line, whose order is defined as the anticlockwise order of the four vertices. The overview of the proposed method is shown in Fig.~\ref{network_overall}. Given an around-view image $\mathbf{I}\in \mathbb{R}^{H\times W\times 3}$ where $H$ and $W$ denote its height and width, the proposed method first uses CNN to extract deep image features. The image features are sent into the marking-point detector to detect marking-points, and the marking-point encoder network to extract the marking-point features (Sec. \ref{network-1}). Given the detected marking-points and corresponding point features, we design attentional graph convolutional neural network to infer the relationship between marking-points (Sec. \ref{network-2}). Our graph is a fully-connected graph, where each node represents a marking-point. We finally apply the entrance line discriminator network to decide whether or not two marking-points form an entrance line (Sec. \ref{network-3}). The whole model is an end-to-end trainable without using manual designed geometric rules.

\subsubsection{Graph Feature Encoder}\label{network-1}
As shown in Fig.~\ref{network_overall}, the image features extracted by the feature extraction network are sent into the marking-point detector and the marking-point feature encoder. The marking-point detector outputs a $S\times S\times 3$ feature map, where the $3$ channels are the marking-point position $x, y$ and confidence $c$ respectively. $N$ marking-points are detected after applying non-maximum suppression (NMS). 

The marking-point feature encoder consists of four convolutional layers and outputs a feature map of size $S\times S\times 64$. With the detected $N$ marking-point positions $\mathbf{P}=(x,y)\in \mathbb{R}^{N\times 2}$, we utilize bilinear interpolation to compute the point-wise features from this feature map. Hence we obtain the marking-point features of size $\mathbf{F}\in \mathbb{R}^{N\times 64}$. Following DMPR-PS \cite{DMPR-PS}, we set the output size $S = 16$ in our implementation.

To enhance the feature representation, we embed the positions of the points $\mathbf{P}$ into a high dimensional vector with Multilayer Perceptron (MLP) and fuse them with the original marking-point features $\mathbf{F}$ via element-wise addition:
\begin{equation} \label{point_encoder}
\mathbf{v}_{i} = \mathbf{f}_{i} + \mathbf{MLP}({x_i, y_i}),
\end{equation}
where $\mathbf{f}_{i}$ is the $i$-th initial marking-point features in $\mathbf{F}$. Equation~\ref{point_encoder} enables the graph network to later reason with joint appearance and position information, which is an instance of the {\em positional encoder} in the Transformer \cite{Transformer}.

\subsubsection{Graph Feature Aggregation}\label{network-2}
Given the above fused marking-point feature $\mathbf{V}$, we build a fully-connected graph $\mathbf{G=(V,E)}$ with $N$ nodes $\mathbf{x}_{i} \in \mathbf{V}$ and $N \times N$ edges $(\mathbf{x}_{i},\mathbf{x}_{j})\in \mathbf{E}$. We then utilize the attentional graph neural network to aggregate the marking-point features. The Graph Neural Network contains several layers. On each layer, it computes an updated representation by simultaneously aggregating messages across all given edges for all nodes. Let $\mathbf{x}_{i}^{l}$ be the feature of the $i$-th node at layer $l$, the feature on layer $l+1$ is represented as :

\begin{equation} \label{equ_all}
\mathbf{x}_{i}^{l+1} = \mathbf{x}_{i}^{l} + \mathbf{MLP}([\mathbf{x}_{i}^{l}\parallel\mathbf{m}_{E_{i}\rightarrow i}^{l}]),
\end{equation}
where $\mathbf{m}_{E_{i}\rightarrow i}^{l}$ is the aggregation result from the first order neighborhood $E_i$ of node $i$ in the graph, and $[\cdot \parallel \cdot]$ represents concatenation. The message aggregation $\mathbf{m}_{E_{i}\rightarrow i}^{l}$ is computed by attention mechanism. Given node feature $\mathbf{x}_{i}^{l}$,  we first compute the query $\mathbf{q}_i$, key $\mathbf{k}_j$ and value $\mathbf{v}_j$ by linear projection:
\begin{equation} \label{attention_coefficients}
	\begin{aligned}
		\mathbf{q}_{i}&=\mathbf{W}_{1}^{l}\mathbf{x}_{i}^{l}+\mathbf{b}_{1}^{l}, \\
		\mathbf{k}_{j}&=\mathbf{W}_{2}^{l}\mathbf{x}_{j}^{l}+\mathbf{b}_{2}^{l}, \\
		\mathbf{v}_{j}&=\mathbf{W}_{3}^{l}\mathbf{x}_{j}^{l}+\mathbf{b}_{3}^{l}, \\
	\end{aligned}
\end{equation}
where $\mathbf{W}_{h}^{l}$ and $\mathbf{b}_{h}^{l}$ are the weight matrix and bias of the $l$-th layer respectively. The attention between node $i$ and $j$ are then computed as the Softmax over the key-query similarities:
\begin{equation} \label{attention_coefficients}
\alpha_{ij} = Softmax(\mathbf{q}_{i}^{T}\mathbf{k}_{j}).
\end{equation}

A single attention head based graph node message aggregation is written by:

\begin{equation} \label{multi-head_attention2}
	\mathbf{m}_{E_i\rightarrow i}^{l} = \sum_{j\in E_i} \alpha _{i,j}^{k} \mathbf{v}_{j}.
\end{equation}

In order to improve model expressivity, multi-head attention mechanism are usually applied, which allows the model to jointly attend to information from different representation subspaces: $\mathbf{m}_{E_i\rightarrow i}^{l} = \mathbf{W}^{l}(\mathbf{m}_{E_i\rightarrow i}^{l,1}\parallel \mathbf{m}_{E_i\rightarrow i}^{l,2} \parallel \cdots \parallel \mathbf{m}_{E_i\rightarrow i}^{l,h})$, where $h$ denotes the number of attention heads, $\mathbf{W}^{l}$ denotes linear transformation weight matrix in the $l$-th layer.

\subsubsection{Entrance Line Discriminator}\label{network-3}
Given the learned node features from the attentional graph neural network, we use an entrance line discriminator to decide whether the marking-point pairs form an entrance line. In particular, two of the node features are concatenated to form a $1\times 128$ input feature, and sent to the line discriminator. The discriminator consists of MLP and dropout layers. The Dropout layers are used only during the training stage. The output is further passed through the sigmoid activation layer to compute the probabilities of forming an entrance line. The final output of the model is a $K\times 5$ matrix, where $K=N\times N$ denotes the number of marking-point pairs. Each of the $K$ vectors consists of 5 elements: $x_1$, $y_1$, $x_2$, $y_2$, $t$, where $(x_1, y_1)$ and $(x_2, y_2)$ denote the position of the parking-slot marking-points and $t$ is the probability that the marking-point pair forms an entrance line. 

\subsubsection{Loss Function}
Losses for both the marking-point prediction and entrance line prediction are considered. We use the mean square error for marking-point prediction and binary cross-entropy loss for entrance line prediction, defined as follows:
\begin{equation} \label{loss1}
\begin{split}
loss = \lambda_1 \cdot loss_{point} + \lambda_2\cdot loss_{line},
\end{split}
\end{equation}
where $loss_{point}$ denotes the marking-point prediction loss, $loss_{line}$ the entrance line prediction loss. $\lambda_1$ and $\lambda_2$ are the weights for balancing the two losses.

As the output layer of the marking-point detector is divided into a grid of $S\times S$ cells, the loss function for the marking-point prediction is defined as the sum of squared errors between the predictions of the $S\times S$ cells and corresponding ground-truths:
\begin{equation} \label{loss2}
\begin{split}
loss_{point} = \frac{1}{S^{2}}\sum_{i=1}^{S^{2}}\{(c_i-\hat{c_i})^2+\mathds{1}_i[(x_i-\hat{x_i})^2+(y_i-\hat{y_i})^2]\},
\end{split}
\end{equation}
where $(x_i, y_i)$ denotes the projected position of marking-point on the grid and $c_i$ is the confidence of cell $i$. Symbols denoted with $\hat{}$ indicate the corresponding ground-truths of the marking-points. $\mathds{1}_i$ equals to $1$ if $(x_i, y_i)$ is the projected ground-truth marking point, otherwise $0$.

The classification loss of the entrance line discriminator is defined as follows:
\begin{equation} \label{loss3}
\begin{split}
loss_{line} = \frac{1}{N^2}(-\sum_{i=1}^{N}\sum_{j=1}^{N}\hat{l_{ij}}logl_{ij}),
\end{split}
\end{equation}
where $l_{ij}$ is the predicted probability of $i$-th and $j$-th marking-point constituting the entrance of parking-slot. Symbols denoted with $\hat{}$ indicate the corresponding ground-truths as above.
\section{Experimental Results and Discussion}
In this section, a series of experiments are conducted to validate the performance of the proposed method.

\subsection{Experiments}
\subsubsection{Datasets}
In this paper, we evaluate the proposed method on the public around-view image dataset called ps2.0 \cite{DeepPS} for vision-based parking-slot detection. The ps2.0 dataset consists of $9827$ training images with $9476$ parking-slots and $2338$ testing images with $2168$ parking-slots. Following DMPR-PS \cite{DMPR-PS}, $7780$ images are selected as the training set and $2290$ images as the testing images. The images in the ps2.0 dataset are collected from typical indoor and outdoor scenes under various environmental conditions, and its resolution is $600 \times 600$ pixels corresponding to a 10m$\times$10m physical plane region. 

\subsubsection{Experimental Setting}

The proposed method is compared with the following state-of-the-art parking-slot detection methods: three traditional parking-slot detection methods: method of Wang et al. \cite{randn-1}, method of Hamda et al. \cite{sobal-2}, and PSD\_L \cite{line}, and two deep-learning based parking-slot detection methods: DeepPS \cite{DeepPS} and DMPR-PS \cite{DMPR-PS}.

Following DeepPS, we adopt the Precision-recall rate as the evaluation metric. $\mathbf{S}_{g}= \left \{\mathbf{p}_{1}^{g},\mathbf{p}_{2}^{g} \right \}$ is set as a ground-truth parking-slot where $\mathbf{p}_{1}^{g}=(x_{1},y_{1})$, $\mathbf{p}_{2}^{g}=(x_{2},y_{2})$ are the corners of the entrance line. $\mathbf{S}_{d}= \left \{\mathbf{p}_{1}^{d},\mathbf{p}_{2}^{d} \right \}$ is a detected marking-point pair. If they satisty the following condition:
\begin{equation} \label{conditions}
\left \|(\mathbf{p}_{1}^{g}-\mathbf{p}_{1}^{d}, \mathbf{p}_{2}^{g}-\mathbf{p}_{2}^{d}) \right \|_2 < 10,
\end{equation}
$\mathbf{S}_{d}$ is regarded as a true positive and $\mathbf{S}_{g}$ is correctly detected. Otherwise, $\mathbf{S}_{d}$ is a false positive and $\mathbf{S}_{g}$ is a false negative parking-slot.

\subsubsection{Implementation Details}
We use Pytorch to implement our method, and the model is trained on Nvidia Titan Xp for $200$ epochs with the Adam optimizer \cite{kingma2014adam}. The initial learning rate is $0.001$ and the batch size is set as $24$. All experiments are performed using one Nvidia Titan Xp GPU. The parameters $\lambda_1$ and $\lambda_2$ are set as $100.0$ and $1.0$ respectively. VGG16 \cite{vgg} is selected as the backbone of image feature extraction. The number of layers and heads of the attentional graph neural network are set as $3$ and $4$ respectively.

\subsection{Results and Discussions}

\begin{table}
	\caption{Precision and recall on the ps2.0 testing set}	
	\begin{center}
		\setlength{\tabcolsep}{6mm}{
		\begin{tabular}{ccc}
			\hline
			method& Precision & Recall\\ 
			\hline
			Wang et al. \cite{randn-1}& 98.29\% & 58.33\% \\ 
			Hamda et al. \cite{sobal-2}& 98.45\% & 61.37\% \\ 
			PSD\_L \cite{line}& 98.41\% & 86.96\% \\ 
			DeepPs \cite{DeepPS}& 98.99\% & 99.13\% \\ 
			DMPR-PS \cite{DMPR-PS}& 99.42\% & 99.37\% \\ 
			\hline
			FCN-baseline& 98.79\% & 98.84\% \\ 
			Ours& {\bf 99.56\%} & {\bf 99.42\%} \\
			\hline
		\end{tabular}}
		\label{Compared}
	\end{center}
\end{table}

\begin{table}
	\caption{Precision and recall on the PSV testing set}	
	\begin{center}
		\setlength{\tabcolsep}{6mm}{
		\begin{tabular}{ccc}
			\hline
			method& Precision & Recall\\ 
			\hline
			DeepPs \cite{DeepPS}& 95.68\% & 87.63\% \\ 
			DMPR-PS \cite{DMPR-PS}& 92.73\% & 81.73\% \\ 
			\hline
			Ours& {\bf 97.05\%} & {\bf 90.70\%} \\
			\hline
		\end{tabular}}
		\label{psv}
	\end{center}
\end{table}

\subsubsection{Evaluation}
Table~\ref{Compared} shows the quantitative results on ps2.0 test set. The proposed method outperforms three traditional parking-slot detection methods with a significant margin, which demonstrates that learning-based approaches can learn more discriminative features, especially for complex visual conditions. Our method achieves comparable performance to the state-of-the-art learning-based method DMPR-PS both in precision and recall. It is noteworthy that DMPR-PS is a two-stage method, and it firstly predicts the position, direction and shape of the marking-points by convolution neural network and then uses manually designed geometric rules to infer the final detected parking-slots. On the contrary, by treating the marking-points as graph-structured data, the proposed method just needs to predict the position of the marking-points and then infers the parking-slot directly with attentional graph neural network in an end-to-end way. Moreover, compared with DMPR-PS, the proposed method does not require fine grained annotations of direction and shape of marking-points, thus reduces the training expenses, which is more suitable to real-world application.

As ps2.0 data has reached its limit for state-of-the-art methods, we use PSV dataset \cite{wu2018vhhfcn} as out-of-distribution testing dataset to verify the generalization performances, {\ie}, we train models on ps2.0 dataset and test them on PSV dataset using the ground truth from \cite{li_vacant_2020}. Table~\ref{psv} lists the detection results of DeepPS, DMPR-PS and our method. Our method outperforms DeepPS by $1.4\%$ on precision rate and $3.1\%$ on recall rate. Compared to DMPR-PS,  our method obtains $4.3\%$ higher precision rate and $9.0\%$ higher recall rate. The results show that our method has good generalization ability for cross-domain parking-slot detection. 

\subsubsection{Effectiveness of Attentional Graph Neural Network}

\begin{figure}
	\centering
	\includegraphics[width=7cm]{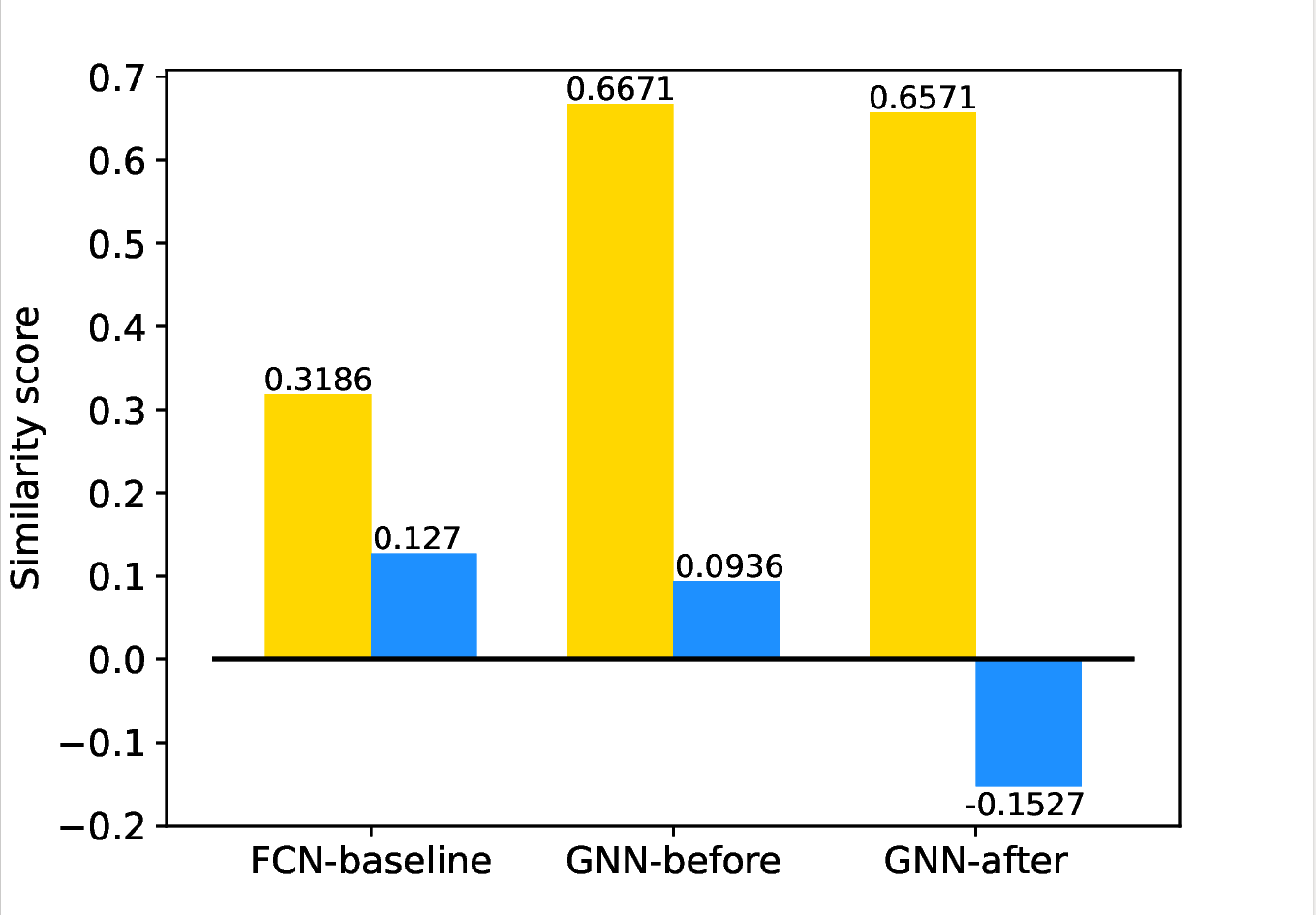}
	\caption{Average similarity score between features of two marking-points. The average similarity score of paired marking-points forming parking-slots is shown in yellow and unpaired marking-points in blue.}
	\label{Similarity}
\end{figure}

\begin{figure*}
	\centering
	\begin{minipage}[b]{.8\linewidth}
		\centerline{\includegraphics[width=14cm]{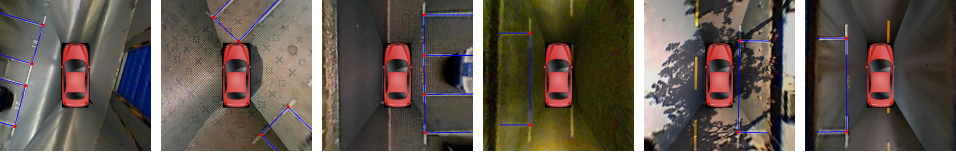}}
	\end{minipage}
	\caption{Qualitative results of parking-slot detection under various conditions, such as indoors and outdoors, daytime and nighttime, sunny and rainy days, etc. Detected Marking-points are drawn in red. Entrance and side lines are drawn in blue.}
	\label{parking}
\end{figure*}

This section studies how the designed attentional graph neural network affects the performance of parking-slot detection. We compare with naive full connection layers to justify the effectiveness of our graph-based method. In particular, we replace the attentional graph neural network with a fully connected neural network, denoted by FCN-baseline. As shown in Table~\ref{Compared}, the proposed method gives a $0.74\%$ higher precision and a $0.58\%$ higher recall rate than FCN-baseline. The reason is that FCN-baseline assumes that data points are independent of each other and ignores each marking-point is related to others. Leveraging such interdependence of different marking-points can be beneficial to accurate parking-slot detection. 

In order to verify how effective the attentional graph neural network can capture the link information of marking-points, we calculate the average cosine similarity between features of all detected marking-points pairs in Fig.~\ref{Similarity}. The average similarity score for all paired marking-points forming parking-slots is shown in yellow while the average similarity score for unpaired marking-points in blue. FCN-baseline, GNN-before and GNN-after denote that we calculate the average cosine similarity between features of two marking-points after the full connection layers, before the attentional graph neural network and after the attentional graph neural network, respectively.  For two paired marking-points, the larger the cosine similarity score the better, while for two unpaired marking-points, the smaller the cosine similarity score the better. From Fig.~\ref{Similarity} we can see that the average cosine similarity between features of two paired marking-points in FCN-baseline is $0.3168$, while the results of the attentional graph neural network are $0.6671$ and $0.6571$, which shows that the features of two paired marking-points in attentional graph neural network are more similar. Compared GNN-before with GNN-after, the features of paired marking-points after the attentional graph neural network are still similar as shown in Fig.~\ref{Similarity}, which indicates that the attentional graph neural network keeps the similarity of the paired marking-points. The similarity between features of unpaired marking-points drops from $0.0936$ to $-0.1527$ after the attentional graph neural network, which verifies that the attentional graph neural network can enlarge the dissimilarity of the unpaired marking-points. We can draw a conclusion that the designed attentional graph neural network has the ability to capture the link information, {\ie}, features of paired marking-points forming parking-slots are similar and features of unpaired marking-points are dissimilar, and thus learn more discriminative features for better parking-slot detection.

We also present several parking-slot detection examples in Fig.~\ref{parking}. As shown, our method provides highly accurate parking-slot detection under various conditions. However, there are also failure cases, especially when marking-points are occluded. This is because our model relays on the detection of the marking-points. How to infer parking-slot with partially observed marking-points is an interesting and challenging problem. We leave this as our future work. The current model is designed for around-view images and it requires annotations of the marking-point positions as well as entrance line information. It may need some extra efforts to work with other types of images, {\eg}, side-view images as in \cite{context:2020}. 

\subsection{Ablation Study}
In this section, extensive ablation studies are performed to validate the strengths of the key components of the proposed approach.
\begin{table}
	\caption{Precision and recall with different backbones}
	\begin{center}
	    \setlength{\tabcolsep}{5mm}{
		\begin{tabular}{cccc}
			\hline
			Backbone& Precision & Recall & Speed (ms)\\ 
			\hline
			ResNet18& 98.54\% & 98.73\% & 16.2 \\ 
			ResNet50& 98.94\% & 98.94\% & 21.4\\ 
			Darknet19 & 98.45\% & 99.07\% & 18.2 \\  
			VGG16& {\bf 99.56\%} & {\bf 99.42\%} & 25.3 \\
			\hline
		\end{tabular}}
		\label{backbone}
	\end{center}
\end{table}

\begin{table}
	\caption{Precision and recall with different variants}
	\begin{center}
		\begin{tabular}{ccccc}
			\hline
			Positional encoder&GNN layer&GNN head& Precision & Recall \\ 
			\hline
			$\times$&3&4 & 99.08\% & 99.42\% \\  
			$\checkmark$&3&2 &98.80\%&{\bf99.51\%} \\  
			$\checkmark$&1&4 & 99.08\% & 99.37\% \\    	
			$\checkmark$&3&4& {\bf 99.56\%} & 99.42\% \\
			\hline
		\end{tabular}
		\label{variants}
	\end{center}
\end{table}

\begin{table}
	\caption{Impacts of loss function weights}
	\begin{center}
		\setlength{\tabcolsep}{7mm}{
		\begin{tabular}{cc|cc}
			\hline
			\multicolumn{2}{c|}{Loss Weight} &\multicolumn{1}{c}{\multirow{2}[0]{*}{Precision}}&\multicolumn{1}{c}{\multirow{2}[0]{*}{Recall}} \\ \cline{1-2} 
			$\lambda_1$ & $\lambda_2$ \\
			\hline
			1&1& 94.69\% & 97.39\% \\ 
			10&1 & 97.90\% & 99.27\% \\  
			100&1 & {\bf 99.56\%} & {\bf 99.42\%} \\
			\hline
		\end{tabular}}
		\label{loss_weight}
	\end{center}
\end{table}

\subsubsection{Different Backbones}
We first study the influence of the backbone for image feature extraction. The precision-recall scores of different backbones are shown in Table~\ref{backbone}. It is shown that the vgg16-based backbone performs better than the other backbones. We also test the inference speed with different backbones on single GTX 1080 Ti GPU. VGG16 backbone takes $25.3$ ms per image, which is around $40$ Hz detection speed, satisfying real-time requirement. The fastest backbone comes with ResNet18, which takes $16.2$ ms per image. The parking-slot detection speed of the multi-state models \cite{context:2020} and \cite{li_vacant_2020} are $58.59$ ms and $42.79$ ms respectively.

\subsubsection{Different Variants}
To evaluate the effectiveness of the proposed model, we do experiments with different model variants. This ablation study, presented in Table~\ref{variants}, shows that all blocks of the proposed model are useful and result in the substantial performance increases. The positional encoder can help to detect more accurately the parking-slots and improves the precision score. The model is not sensitive to the configurations of the attentional graph neural network. The empirical study shows that with three layers and four heads, the proposed model performs best.

\subsubsection{Loss Weight}
We conduct experiments with various combinations of weights of marking-point prediction loss and entrance line prediction loss. As it is shown in Table~\ref{loss_weight}, the proposed model prefers a larger loss weight at the marking-point prediction. This is because that the task of marking-point prediction is more difficult than the task of entrance line prediction.

\section{Conclusion}
In this paper, we propose an attentional graph neural network based model for parking-slot detection. The proposed method takes a single around-view image as input, and detects the  parking-slots in an end-to-end manual without any post-processing. The core contribution of our method is the utilizing of the graph neural network to mine the relationship of marking-points, which bridges the detection of marking-points and location of parking-slots. Experimental results on ps2.0 and PSV dataset have shown that our method achieves comparable accuracies to the state-of-the-art CNN-based method DMPR-PS. In the future, we are planning to extend our method to more complex parking-slot detection scenarios, such as oblique, trapezoid and stereo parking-slots.

\bibliographystyle{IEEEtran}
\bibliography{references}

\begin{thebibliography}{10}
\providecommand{\url}[1]{#1}
\csname url@samestyle\endcsname
\providecommand{\newblock}{\relax}
\providecommand{\bibinfo}[2]{#2}
\providecommand{\BIBentrySTDinterwordspacing}{\spaceskip=0pt\relax}
\providecommand{\BIBentryALTinterwordstretchfactor}{4}
\providecommand{\BIBentryALTinterwordspacing}{\spaceskip=\fontdimen2\font plus
\BIBentryALTinterwordstretchfactor\fontdimen3\font minus
  \fontdimen4\font\relax}
\providecommand{\BIBforeignlanguage}[2]{{%
\expandafter\ifx\csname l@#1\endcsname\relax
\typeout{** WARNING: IEEEtran.bst: No hyphenation pattern has been}%
\typeout{** loaded for the language `#1'. Using the pattern for}%
\typeout{** the default language instead.}%
\else
\language=\csname l@#1\endcsname
\fi
#2}}
\providecommand{\BIBdecl}{\relax}
\BIBdecl

\bibitem{DeepPS}
M.~Heimberger, J.~Horgan, C.~Hughes, J.~Mcdonald, and S.~Yogamani, ``Computer
  vision in automated parking systems: Design,implementation and challenges,''
  \emph{Image and Vision Computing}, vol.~68, no. dec., pp. 88--101, 2017.

\bibitem{PS-DCNN}
W.~Li, H.~Cao, J.~Liao, J.~Xia, L.~Cao, and A.~Knoll, ``Parking slot detection
  on around view images using dcnn,'' \emph{Frontiers in Neurorobotics},
  vol.~59, no.~2, pp. 616--626, 2010.

\bibitem{2000-first}
J.~Xu, G.~Chen, and M.~Xie, ``Vision-guided automatic parking for smart car,''
  in \emph{\IV}, 2002.

\bibitem{2006-second}
H.~G. Jung, D.~S. Kim, P.~J. Yoon, and J.~Kim, ``Structure analysis based
  parking slot marking recognition for semi-automatic parking system,'' in
  \emph{IAPR Int. Workshop Struct. Syntact. Patt. Recog.}, 2006.

\bibitem{2010-third}
H.~G. Jung, Y.~H. Lee, and J.~Kim, ``Uniform user interface for semiautomatic
  parking slot marking recognition,'' \emph{IEEE Transactions on Vehicular
  Technology}, vol.~59, no.~2, pp. 616--626, 2010.

\bibitem{DMPR-PS}
J.~Huang, L.~Zhang, Y.~Shen, H.~Zhang, and Y.~Yang, ``Dmpr-ps: A novel approach
  for parking-slot detection using directional marking-point regression,'' in
  \emph{IEEE International Conference on Multimedia and Expo (ICME)}, 2019.

\bibitem{GCN}
T.~N. Kipf and M.~Welling, ``Semi-supervised classification with graph
  convolutional networks,'' in \emph{\ICLR}, 2017, pp. 1--14.

\bibitem{Transformer}
A.~Vaswani, N.~Shazeer, N.~Parmar, J.~Uszkoreit, L.~Jones, A.~N. Gomez,
  L.~Kaiser, and I.~Polosukhin, ``Attention is all you need,'' in \emph{\NIPS},
  2017, p. 6000–6010.

\bibitem{randn-1}
C.~Wang, H.~Zhang, M.~Yang, X.~Wang, L.~Ye, and C.~Guo, ``Automatic parking
  based on a bird's eye view vision system,'' \emph{Advances in Mechanical
  Engineering}, vol. 2014, pp. 847\,406--847\,406, 2014.

\bibitem{sobal-1}
H.~G. Jung, D.~S. Kim, P.~J. Yoon, and J.~Kim, ``Parking slot markings
  recognition for automatic parking assist system,'' in \emph{\IV}, 2006.

\bibitem{harris-1}
J.~K. Suhr and H.~G. Jung, ``Full-automatic recognition of various parking slot
  markings using a hierarchical tree structure,'' \emph{Optical Engineering},
  vol.~52, no.~3, p. 7203, 2013.

\bibitem{harris-2}
J.~Suhr and H.~Jung, ``Sensor fusion-based vacant parking slot detection and
  tracking,'' \emph{\TITS}, vol.~15, no.~1, pp. 21--36, 2014.

\bibitem{line}
L.~Li, L.~Zhang, X.~Li, X.~Liu, and L.~Xiong, ``Vision-based parking-slot
  detection: A benchmark and a learning-based approach,'' in \emph{IEEE
  International Conference on Multimedia and Expo (ICME)}, 2017.

\bibitem{gcn:2009}
F.~{Scarselli}, M.~{Gori}, A.~C. {Tsoi}, M.~{Hagenbuchner}, and
  G.~{Monfardini}, ``The graph neural network model,'' \emph{IEEE Transactions
  on Neural Networks}, vol.~20, no.~1, pp. 61--80, 2009.

\bibitem{bruna2013spectral}
J.~Bruna, W.~Zaremba, A.~Szlam, and Y.~Lecun, ``Spectral networks and locally
  connected networks on graphs,'' in \emph{\ICLR}, 2014, pp. 1--14.

\bibitem{defferrard2016convolutional}
M.~Defferrard, X.~Bresson, and P.~Vandergheynst, ``Convolutional neural
  networks on graphs with fast localized spectral filtering,'' in \emph{\NIPS},
  2016, pp. 3844--3852.

\bibitem{GAT}
P.~Velikovi, G.~Cucurull, A.~Casanova, A.~Romero, P.~Liò, and Y.~Bengio,
  ``Graph attention networks,'' in \emph{\ICLR}, 2018, pp. 1--12.

\bibitem{gcn_pose_estimation}
S.~Jin, W.~Liu, E.~Xie, W.~Wang, C.~Qian, W.~Ouyang, and P.~Luo,
  ``Differentiable hierarchical graph grouping for multi-person pose
  estimation,'' in \emph{\ECCV}, 2020, pp. 718--734.

\bibitem{Superglue}
P.-E. Sarlin, D.~DeTone, T.~Malisiewicz, and A.~Rabinovich, ``Superglue:
  Learning feature matching with graph neural networks,'' in \emph{Proceedings
  of the IEEE/CVF Conference on Computer Vision and Pattern Recognition}, 2020,
  pp. 4938--4947.

\bibitem{poitcloud-grpah}
Y.~Wang, Y.~Sun, Z.~Liu, S.~E. Sarma, M.~M. Bronstein, and J.~M. Solomon,
  ``Dynamic graph cnn for learning on point clouds,'' \emph{Acm Transactions on
  Graphics}, vol.~38, no.~5, 2018.

\bibitem{sobal-2}
K.~Hamada, Z.~Hu, M.~Fan, and H.~Chen, ``Surround view based parking lot
  detection and tracking,'' in \emph{\IV}, 2015.

\bibitem{kingma2014adam}
D.~Kingma and J.~Ba, ``Adam: A method for stochastic optimization,'' in
  \emph{\ICLR}, 2014, pp. 1--15.

\bibitem{vgg}
K.~Simonyan and A.~Zisserman, ``Very deep convolutional networks for
  large-scale image recognition,'' in \emph{\ICLR}, 2015, pp. 1--14.

\bibitem{wu2018vhhfcn}
Y.~Wu, T.~Yang, J.~Zhao, L.~Guan, and W.~Jiang, ``Vh-hfcn based parking slot
  and lane markings segmentation on panoramic surround view,'' in \emph{\IV},
  2018, pp. 1767--1772.

\bibitem{li_vacant_2020}
W.~Li, L.~Cao, L.~Yan, C.~Li, X.~Feng, and P.~Zhao, ``Vacant parking slot
  detection in the around view image based on deep learning,'' \emph{Sensors},
  vol.~20, pp. 2138--2146, 2020.

\bibitem{context:2020}
H.~{Do} and J.~Y. {Choi}, ``Context-based parking slot detection with a
  realistic dataset,'' \emph{IEEE Access}, vol.~8, pp. 171\,551--171\,559,
  2020.

\end{thebibliography}

\end{document}